\documentclass[10pt,journal,compsoc]{IEEEtran}
\usepackage[utf8]{inputenc}
\usepackage{fullpage}
\usepackage{subfigure}
\usepackage{graphicx}
\usepackage{verbatim}
\usepackage{hyperref}
\usepackage{enumitem}
\usepackage{todonotes}  

\usepackage{listings}
\usepackage{soul}
\usepackage{multibib}
\usepackage{draftwatermark}
\SetWatermarkText{Preprint, Accepted by IEEE Computer}
\SetWatermarkScale{0.27}

\title{See the World through Network Cameras}
\author{
Yung-Hsiang Lu,
George K. Thiruvathukal,
Ahmed S. Kaseb,
Kent Gauen,
Damini Rijhwani,\\
Ryan Dailey,
Deeptanshu	Malik,
Yutong	Huang,
Sarah Aghajanzadeh,
Minghao Guo
}

\date{}

\begin{document}
\maketitle

\begin{abstract}
Millions of network cameras have been deployed worldwide. Real-time
data from many network cameras can offer instant views of multiple locations
with applications in public safety, transportation management, urban
planning, agriculture, forestry, social sciences, atmospheric
information, and more. This paper describes the real-time data
available from worldwide network cameras and potential applications.
Second, this paper outlines the CAM$^2$ System available to users at
\url{https://www.cam2project.net/}.  This information includes
strategies to discover network cameras and create the camera database, user
interface, and computing platforms.  Third, this paper describes many
opportunities provided by data from network cameras and challenges to
be addressed.
\end{abstract}


\section{Overview}

\begin{figure*}
\centering
\subfigure[]
{\includegraphics[height=0.95in]{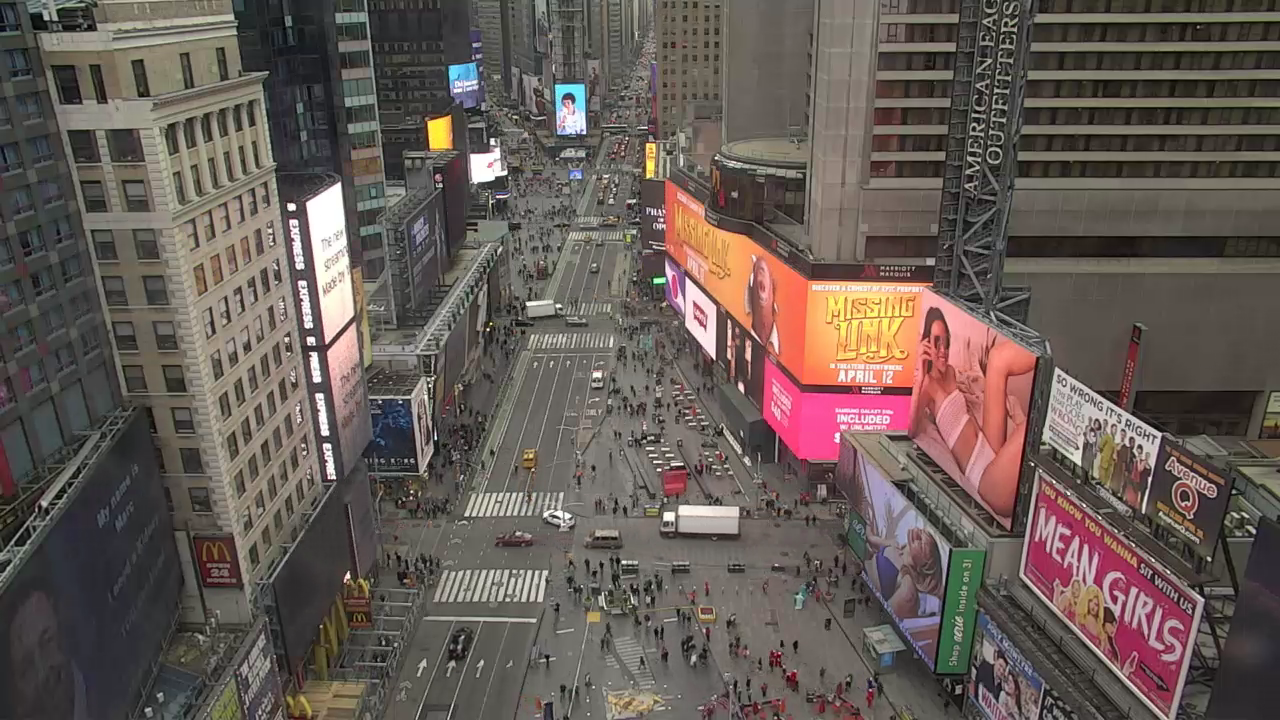}}
\subfigure[]
{\includegraphics[height=0.95in]{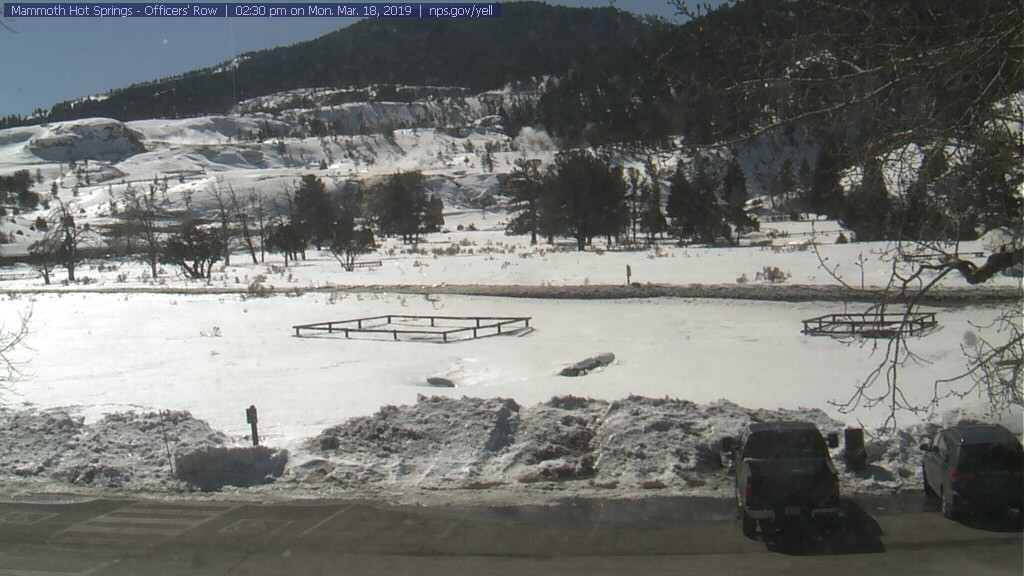}}
\subfigure[]
{\includegraphics[height=0.95in]{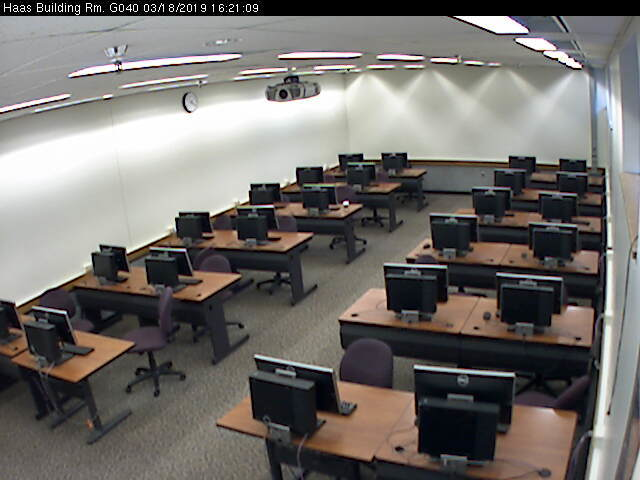}}
\subfigure[]
{\includegraphics[height=0.95in]{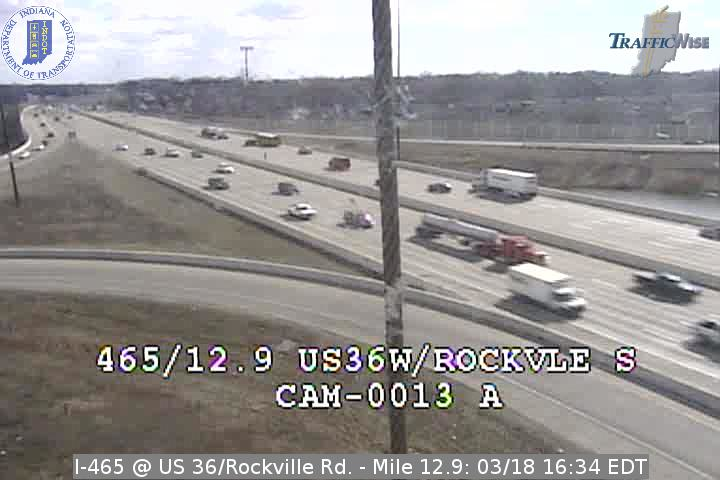}}
\subfigure[]
{\includegraphics[height=1.4in]{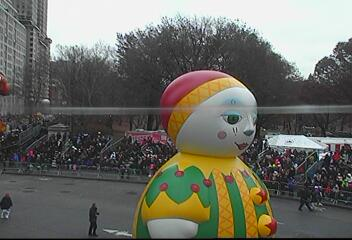}}
\subfigure[]
{\includegraphics[height=1.4in]{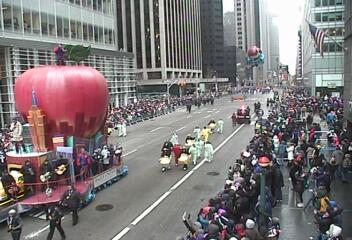}}
\subfigure[]
{\includegraphics[height=1.4in]{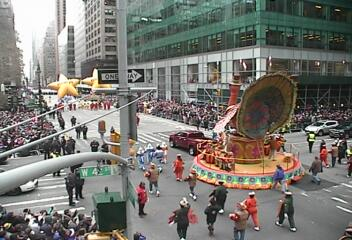}}
\caption{(a) New York City. (b) Yellowstone National Park. (c) A computer lab. (d) 
I-465 highway in Indianapolis. (e)-(g) Thanksgiving Day Parade in New York City. All visual data in this paper is obtained from the Internet and publicly available without
a password.}
\label{fig:sample_images}
\end{figure*}

The first network camera was, perhaps, deployed at the University of Cambridge in 1993 for watching
a coffee pot~\cite{Kesby2012Howworldsfirst}.   Millions of stationary cameras (also called surveillance cameras or webcams in some cases) have been installed at traffic intersections, laboratories, shopping malls, national parks, zoos, construction sites, airports, country borders, university campuses, classrooms,  building entrances, etc. These network cameras can provide  visual data (image or video) continuously without human intervention.  The data from some (but not all) cameras are recorded for post-event (also called forensic) analysis.
This paper explores the opportunities for analyzing  data streams from thousands of network cameras simultaneously. Real-time data may be used in emergency responses; archival data may be used
for discovering long-term trends.
Figure~\ref{fig:sample_images} shows several examples of visual data from
network cameras. As can be seen in these examples, the content varies widely from indoor
to outdoor and urban to natural environments. 
This paper 
considers analyzing the data from many heterogeneous network cameras in real-time.  The paper describes:
\begin{itemize}[noitemsep,nolistsep]
\item Section~\ref{sec:applications}: Potential applications for real-time network camera data
\item Section~\ref{sec:cam2}: The Purdue CAM$^2$ System, including the (1) discovery and retrieval system for network cameras, (2) metadata collection details, (3) web user interface to visualize the locations of the cameras and recent snapshots, (4) system architecture to support real-time data analysis, and (5) resource manager to scale computational resources based on needs
\item Section~\ref{sec:opp}: The opportunities and challenges faced when utilizing data from network cameras
\end{itemize}

\vspace{0.2in}
\begin{tabular}{|p{3in}|} \hline
\begin{center}
{\bf Network Cameras}
\end{center}

There is no universally accepted definition of {\it network cameras}. This paper adopts the following definition: a network camera is connected to a network (the Internet or intranet) and can capture visual data automatically 
and indefinitely without human effort. 
A network camera may have movement (or pan-tilt-zoom, PTZ) capability.
The cameras may send video streams continuously, take periodic snapshots, or acquire data when events are triggered (such as motion detection). 
Most network cameras are stationary; i.e., their locations are fixed.
It is also possible to have mobile network cameras; some cruise ships
take periodic snapshots of oceans and transmit the data through satellite
networks. Some dashcams have network interfaces and may transmit data
while the vehicles are moving or parked.

\\
\hline
\end{tabular}
\vspace{0.2in}

\section{Potential Applications}\label{sec:applications}

Analyzing visual data (image or video) has been an active research
topic for decades. Historically, researchers analyze the data taken in
laboratories.  In recent years, media hosting services (such as
Flickr, Youtube, and Facebook) make sharing visual data much
easier. Researchers start using the data acquired from the Internet to
create datasets, such as ImageNet~\cite{ImageNet} and COCO (Common
Objects in Context)~\cite{COCO}. Most studies are ``off-line'': the
analysis is conducted long after the data has been acquired and there
is no specific restriction on the execution time.  Often, only pixels
are available and there is no time or location information about the
data. As a result, it is not possible to link the data with the
"context", such as breaking news or a scheduled event.  Furthermore,
these datasets do not differentiate data taken from city downtowns or
national parks. One exception uses periodic snapshots to observe
seasonal trends in
environments~\cite{OSullivan2014Democratizingvisualization50015}; the
study considers low refresh rates (a few images from each camera per
day).  In contrast, this paper considers data at much higher refresh
rates (video or snapshots every few minutes).  Adding time and
location information can have profound impacts on how the data can be
used, as explained in the following examples.

\subsection{Virtual Tour}

The world bank estimates international tourists reached 1.2 billion in
2015.  
Nothing can replace the personal experience of visiting a place,
enjoying the culture and the local food; however, the hassle of
traveling can be unpleasant. Many tourist attractions install network
cameras and provide real-time data, such as the Yellowstone National
Park and the National Zoo shown in Figures~\ref{fig:sample_images} (b).  Through these cameras, it is possible to provide ``virtual
tours'' to visitors.  Moreover, it is also possible using network
cameras to watch scheduled events. Figures~\ref{fig:sample_images}
(e)-(g) show images taken in New York City during the Thanksgiving
Day Parade in 2014.

\subsection{Air Quality}

The National Park Service (NPS) deploys network cameras monitoring air
quality~\cite{NPSAirQualityWeb}. Each camera takes one image every 15
minutes and posts the image on the NPS web site.  The data is archived
and can be used to study phenology.  The data can be cross referenced
with other sources of data such as the archive of weather data
(humidity, temperature, cloudiness) and rare events such as
wildfire. In addition to these cameras deployed in national parks,
many TV stations deploy cameras watching cities. These network cameras
may also be used to assess the air quality in the cities.

\subsection{Transportation Management and Urban Planning}

\begin{figure*}
\centering
\subfigure[]
{\includegraphics[height=1.9in]{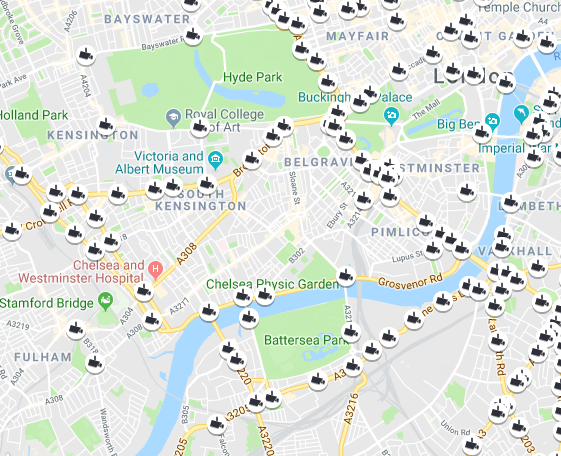}}
\hspace{0.1in}
\subfigure[]
{\includegraphics[height=1.9in]{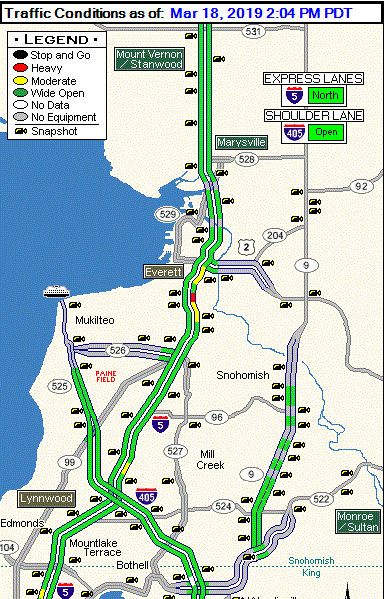}}
\hspace{0.1in}
\subfigure[]
{\includegraphics[height=1.85in]{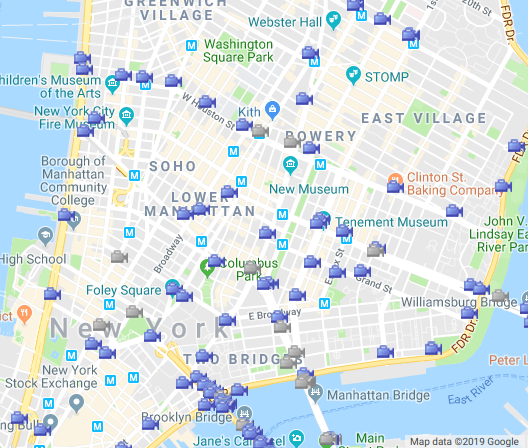}}
\subfigure[]
{\includegraphics[height=1.85in]{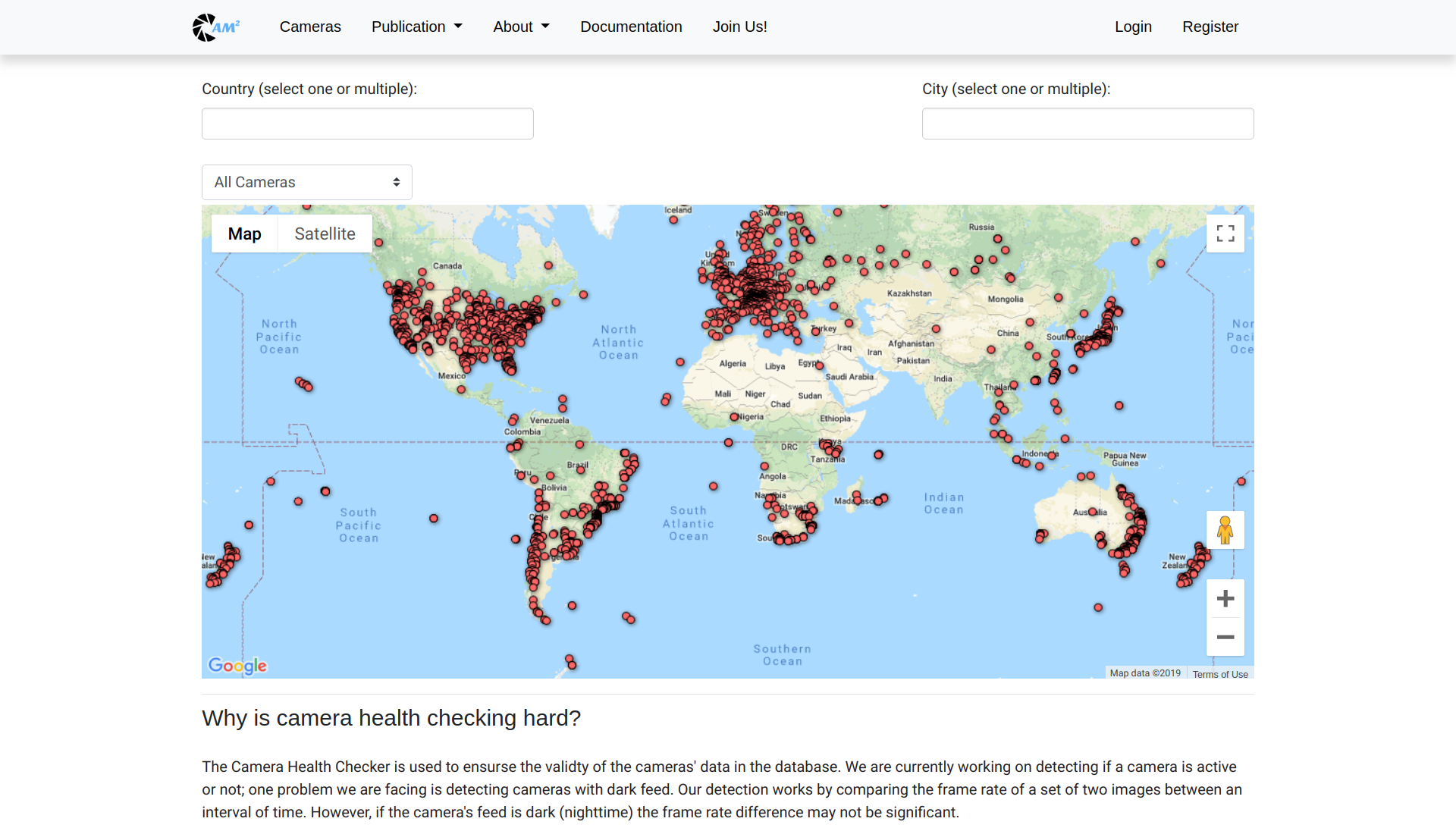}}
\subfigure[]
{\includegraphics[height=1.8in]{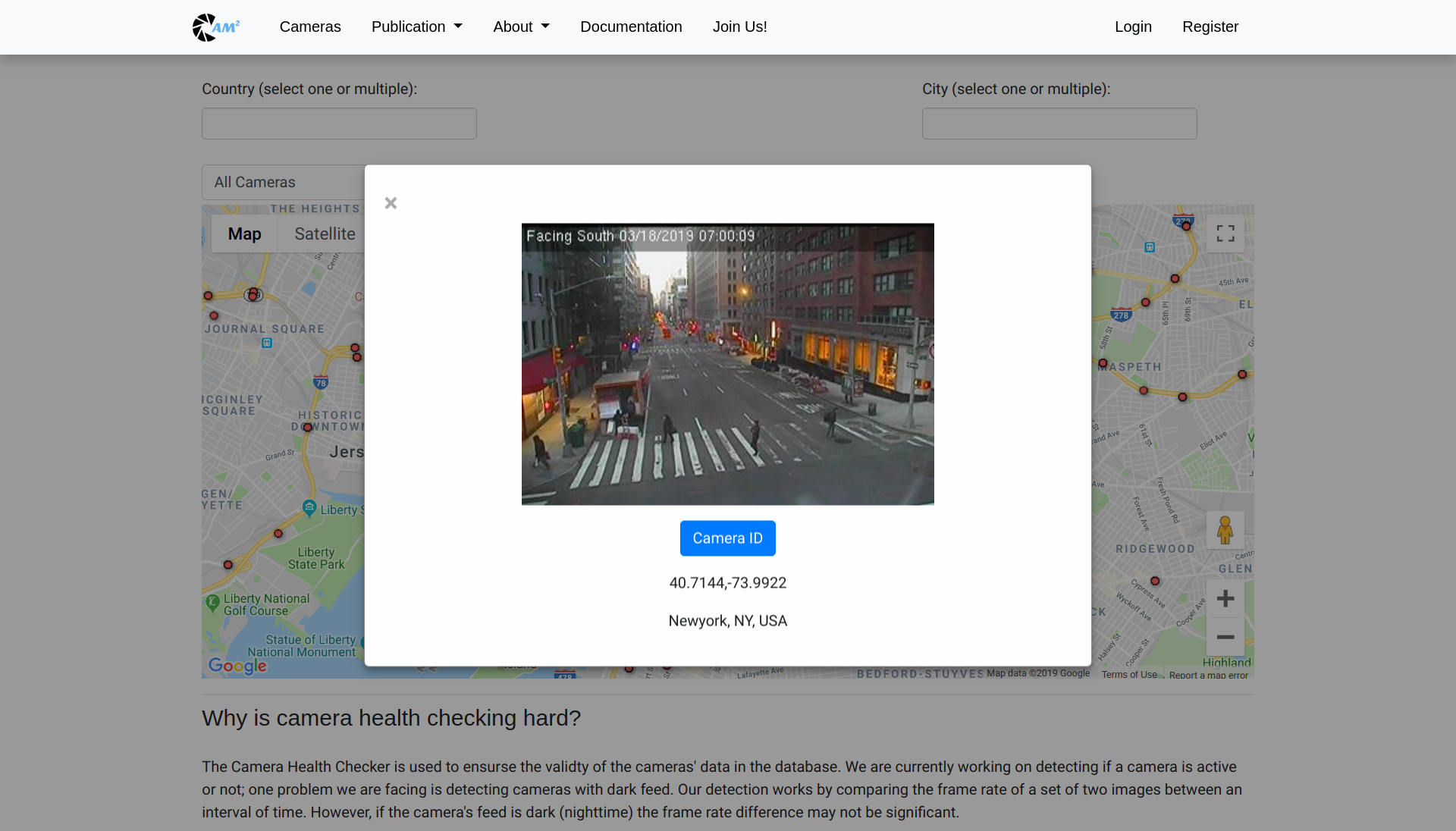}}
\caption{Maps of traffic cameras in (a) London, (b) Seattle, (c) New York. Screenshots from the CAM$^2$ website (d) worldwide camera location map and (e) New York camera location map with one real-time image.}
\label{fig:trafficcameras}
\end{figure*}

Improving transportation efficiency is a significant challenge in many
cities (Chicago, Houston, London, Seattle, New York, etc.). Network
cameras are widely deployed at traffic intersections.  Currently, the
real-time data allows city officials to monitor traffic congestion. In the
future, this processes could be automatically optimized based on the
real-time traffic information provided by the network
cameras. Figure~\ref{fig:sample_images}~(d) is an example of
a traffic camera in Indianapolis.
Figures~\ref{fig:trafficcameras} (a)-(c) show the locations of traffic cameras
in London, Seattle, and New York.

\subsection{Safety and Emergency Response}

\begin{figure}
\centering
\subfigure[]
{\includegraphics[height=1.2in]{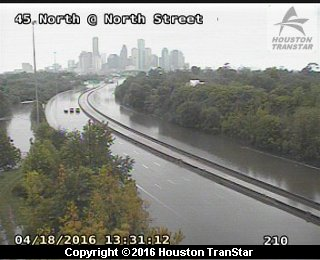}}
\subfigure[]
{\includegraphics[height=1.2in]{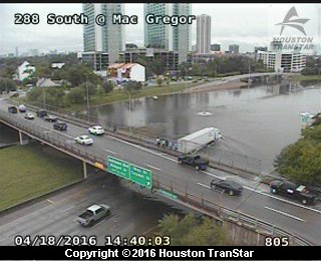}}
\subfigure[]
{\includegraphics[height=1.2in]{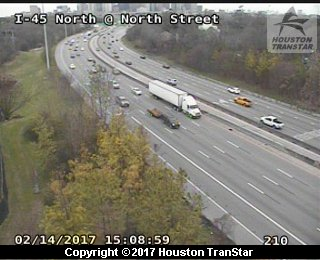}}
\subfigure[]
{\includegraphics[height=1.2in]{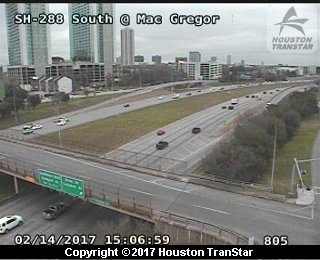}}
\caption{(a)(b) Houston Flood on 2016/04/18. (c)(d) Normal condition on 2017/02/14 taken by the same cameras.}
\label{fig:houston}
\end{figure}

It is possible using network cameras to monitor large-scale emergencies.
Figures~\ref{fig:houston} (a) and (b) show the flood in Houston on 2016/04/18.
Since network cameras continuously acquire data, it is possible to conduct
``before-after'' comparison as shown in
Figures~\ref{fig:houston} (c) and (d) when the highways returned to the normal conditions.
Our recent study~\cite{Surakitbanharn2018HST} suggests that data from network
cameras can complement postings on social networks during emergencies. Network cameras
continuously acquire and transmit data without human efforts; thus, network cameras 
can be used to monitor locations that have already been evacuated.

\begin{figure*}[h]
\centering
\includegraphics[width=6.5in]{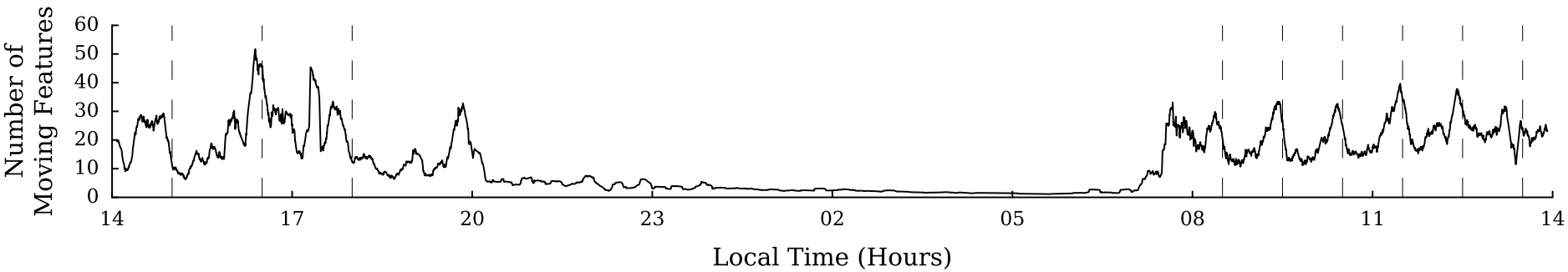}
\caption{Tracking moving features in a video stream of a camera at Purdue University. More moving features exist during the day, especially before the starting times of lectures (indicated by the vertical dashed lines).}
\label{fig:experiment24}
\end{figure*}

\subsection{Human Activities}

An experiment  tracks the moving features in a video stream from a camera at Purdue University for 24 hours~\cite{KasebTCC}. 
The experiment analyzes 820,000 images (approximately 10 frames per second, FPS) from a single camera.
Figure~\ref{fig:experiment24} shows that more moving features are detected during the day, especially before the lectures' starting times.
This experiment demonstrates that it is possible to gain insights about the behavior of people using relatively simple analysis programs.

\subsection{Versatile Data from Network Cameras}

Computer vision has made significant progress in recent years. One
factor contributing to the success is large datasets with thousands or
millions of images and labels.  Different datasets may have specific
emphasis~\cite{GauenIRI}.  For example, images
posted on social networks tend to have faces at or near the images'
centers.  Video captured by dashcams tend to have pedestrians at the
horizon. Traffic cameras are usually 3-stories high looking
downwards. These characteristics are a result of the sampling images
from different data distributions. The difference among different
datasets can be called ``distinctiveness''. Distinctiveness can be
desirable because datasets focus on specific purposes---for face
recognition, data from traffic cameras may not be useful.  Data from
network cameras provide a wide variety and is a rich source for data
that is not always easily available in research laboratories.

\section{Continuous Analysis of Many CAMeras (CAM$^2$) Project at Purdue
University}\label{sec:cam2}

The previous section describes many examples where analyzing the data
from network cameras (real-time images or video streams) can be
helpful.  This section describes a research project at Purdue
University, CAM$^2$, to construct a system to continuously analyze
visual data from many network cameras. Specifically, this section
outlines how to discover network cameras from many different sources,
retrieve data and metadata from them, the backend required for
analyzing data in real-time, and a close inspection of the
resource manager required for scaling computational needs of analysis
programs.

\subsection{Discover Network Cameras}\label{subsec:discovery}

\vspace{0.1in}
\begin{tabular}{|p{3in}|} \hline
\begin{center}
{\bf IP and Non-IP Cameras}

\end{center}

Many network cameras can be connected to the Internet directly and have unique IP (Internet Protocol) addresses. They are called
``IP-cameras'' in this paper. 
Some cameras (such as webcams) are connected to computers that make data available on the Internet. They
are called ``non-IP cameras'' in this paper because the cameras themselves do not have own IP addresses. Each network camera may have an IP address but does not necessarily expose itself and may rely on a computer to act as a proxy. In this case, the IP address is the computer's IP address, 
not the camera's IP. Many organizations have web servers that
show the data from multiple cameras. Since the IP addresses are the web servers' addresses, 
these cameras are also considered as non-IP cameras.
 \\ \\
\hline
\end{tabular}
\vspace{0.1in}

The applications described above require data from many geographically
distributed network cameras.  The procedure of finding network cameras
and aggregation websites can be found
in~\cite{Dailey2017CreatingWorldsLargest512}. This article summarizes
the process.  IP-cameras usually have built-in web servers and the
data can be viewed through web browsers.  These cameras support HTTP
(Hypertext Transfer Protocol). Different brands have different paths
for retrieving data using the GET commands.  Several methods can be
used to find IP-cameras.  One obvious method queries search
engines. This method, however, has a low success rate because search
engines usually return vendors of network cameras, not IP addresses of
network cameras that can provide real-time data streams. Another
method scans IP addresses by sending the GET commands of the known
brands. If an IP address responds to the commands positively (``200
OK''), then the IP address becomes a candidate network camera.  The
candidate is further inspected by the Purdue team.  Currently, this
process is manual for two reasons. First, some IP addresses respond to
the GET commands even though they are not network cameras (false
positive). Second, the Purdue team inspects the discovered camera and
keeps it only if the camera data is from a public location (such as a
traffic intersection, a park, or a university campus).  CAM$^2$ is
actively investigating the automation of discovering network cameras
(c.f.  Section~\ref{subsec:uniformAPI}). To automate privacy filtering
in the future, we anticipate deep learning models
may become capable of scene classification of private versus public
locations.

\subsection{Metadata Aggregation}\label{subsec:metadata}

Following network camera discovery, collecting additional information
(called ``metadata'') about the cameras is important (and possibly
required) for data analysis.  In this project, metadata includes (but
is not limited to) the cameras' locations, methods to retrieve data,
the format of the data (such as MP4, flash, MJPEG, JPEG, and PNG), and
the information about the refresh rate of the network
cameras. Metadata may also describe the data's content, such as
indoor/outdoor, highway, parks, university campus, shopping malls,
etc.  Three particularly important pieces of metadata are location,
data quality, and data reliability. These are explained in the
following paragraphs.

Location information is required for many applications described
earlier.  In many cases, the owner of a camera provides the precise
location (with longitude and latitude) of the camera. In some other
cases, street addresses are given.  It is also possible to use the IP
addresses to determine the geographic locations but this approach may
be inaccurate for several reasons. An organization (such as a
university) may have a large campus. Knowing that an IP address
belongs to this organization may not provide sufficient details about
the camera's location.  Moreover, as mentioned above, some
organizations show multiple data streams on web sites. The web
servers' locations do not reflect the cameras' locations.  In the
future, accurate locations may be estimated by cross-referencing
information from the network camera images with other resources. Some
examples include (1) the time of day given the sunlight, the direction
and length of shadow~\cite{10.1007/978-3-540-88682-2_25},
(2) current events (like parades), and (3) identifying significant
landmarks.

Data quality is critical for analysis. Data quality can be measured by
many metrics. One is the resolution (number of pixels); another is the
refresh rate (frames per second). The data quality may also be
determined by the purpose of the applications: for example, if a
camera is deployed to monitor traffic, then the data quality is
determined by whether it can see congestion clearly or the view is
blocked by trees. In contrast, if a camera is deployed to monitor air
quality, it is more important to evaluate whether the view has high
visibility.

Reliability refers to the availability of the network camera data. For
example, some network cameras only provide data during the daylight
hours and do not provide data during night-time hours. Some network
cameras are available 24 hours. Some others may be disconnected for
various reasons, such as being damaged during a hurricane.  

\subsection{Web User Interface}\label{subsec:webUI}

CAM$^{2}$ is designed as an open research tool, available for other
researchers to use. Thus, it has a web interface
(\url{https://www.cam2project.net/}) for users to select cameras based
on locations.  Figure~\ref{fig:trafficcameras} (d) is a screenshot of
the website. The locations of the cameras are shown as markers on a
map (using Google Maps). When a marker is clicked, a snapshot is
displayed, as shown in Figure~\ref{fig:trafficcameras} (e).  The web
site allows users to select cameras based on country, state (for
USA.), and city. The map automatically zooms into the selected
country.  The markers in Figure~\ref{fig:trafficcameras} (d) were originally
implemented using the Google Maps client API. However, as the number
of cameras in CAM$^2$ grows, this is no longer a scalable solution.
Experiments showed that loading 10,000 markers would take nearly 20
seconds for rendering the map.  To improve scalability, the CAM$^2$
website uses Google Fusion Tables. This supports tile-based rendering
of the markers on Google Map.  The rendering time for 100,000 markers
is less than 2.5 seconds.

\subsection{System Architecture}\label{subsec:systemArchitecture}

\begin{figure}[hbtp]
  \centering

  \includegraphics[width=3in]{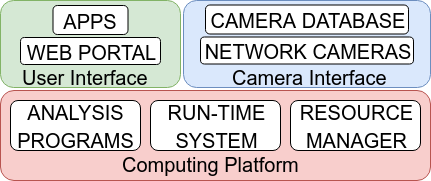}
  \caption{CAM$^2$ has three primary components: User Interface, Camera Interface, and Computing Platform.}
  \label{fig:architecture}
\end{figure}

Figure~\ref{fig:architecture} shows the three primary components of
CAM$^2$~\cite{MakingCAM2}: the user interface, camera interface, and
computing platform. The user interface is made up of two access
points. First, applications (Apps) can be programmed with our Python
API (application programming interface)~\cite{CAM2_System_GlobalSIP}
to access the CAM$^{2}$ system. Second, the user can access CAM$^2$
through a web portal. Aside from the user interaction, the entire
CAM$^2$ system is automated. The web portal allows users to select the camera
data streams for analysis, specify the desired analysis
parameters (e.g., frame rate and duration), and submit the analysis
programs. In other words, the web portal grants users access to the
other two essential features of CAM$^2$.

The camera interface is accessed through the user
interface. The camera database provides access to the network
cameras. It is an SQL database storing the URL of the network cameras
along with other metadata information.  The network cameras themselves
are, of course, already deployed globally.  After or during data
collection, the user can run an analysis program from the computing
platform. For example, an analysis program can be used to track
people (c.f. Figure~\ref{fig:experiment24}).

The computing platform contains three major components, all of which
enables users to run analysis programs on cloud
servers.~\cite{CAM2_System_GlobalSIP}. (1) Analysis programs are
either created by users or selected from an available list of provided
programs.  (2) The run-time system is an event-driven system for processing
images. After a new image (or video frame) is acquired, a call-back function
is invoked.  Currently, the data streams are treated independently;
thus, this system is intrinsically parallel and can scale up to
process thousands of data stream simultaneously. (3) The resource
manager allocates the appropriate number of cloud instances to execute
the analysis programs.  The cloud instances are responsible for
retrieving the visual data from the cameras and executing the analysis
programs in real-time.  The instances may include GPUs (graphics
processing unit).

\subsection{Resource Manager}

Inside the computing platform, the resource manager is a crucial
component of CAM$^2$ for automatically scaling the computational
resources to meet analysis programs' demands.  Some applications
(such as transportation management and emergency response) need to
analyze data only at certain time periods (rush hours or when a
disaster occurs). Thus, the resource manager needs to adjust the
allocated resources as the needs rise and fall. Many factors can
affect the resource manager's decisions.  Cloud vendors offer dozens
of instance types with various amounts of available processor
cores, memory, GPU cores, storage, etc.  Furthermore, cloud instances
of the same capability (same number of cores and same amount of
memory) have up to 40\% of difference in cost~\cite{KasebTCC}.

When the required computation and monetary costs are known for an
analysis program, the optimal solution can be determined via a convex
optimization problem~\cite{7450565}. It assumes computation and memory
use scales linearly with the number of cloud instances; this is a
reasonable assumption since this is the guarentee provided by the host
of a cloud instance.  The paper shows the optimal cloud instance is
the minimum ratio between the cost of a given cloud instance and
the provided computation power (in terms of memory and CPU speed).

To make the problem even more challenging, resource requirements
depend on the content of the data as well as the analysis programs.
A study suggests using multi-dimensional bin packing to
model the relationships between the needs of analysis programs and the
characteristics of cloud instances~\cite{KasebTCC}.  The method
reduces overall cost by up to $60$\%. 

However, when the geographical distance (hence, network round-trip
time) increases, the data refresh rate may decline~\cite{ChenCloud,
8359122}.  The network camera's image quality can suffer.  As a
result, it is necessary to select a data center that is close to the
network cameras if a high refresh rate is desired.  This is an issue
as network cameras are deployed worldwide and cloud data centers are
located in many different parts of the world. Therefore the cost,
location, and required image quality for analysis must be considered
together for determining the proper cloud
instance~\cite{AnupCloudCom}. By modifying the original bin packing
method~\cite{KasebTCC}, the new study shows a reduction in cost by
$56$\% when compared with selecting the nearest location and
further improved the original method by $36$\%.

\section{Opportunities and Challenges}\label{sec:opp}

To realize the applications outlined in Section~\ref{sec:applications}, 
the following research opportunities and challenges must be investigated.

\subsection{Automatically Adding Network Cameras}\label{subsec:uniformAPI}

Adding network cameras to the CAM$^{2}$ database must be further
automated to utilize the vast amount of network camera data still yet
to be discovered. The challenges of using public network camera data
leave this valuable data source largely unused. Network camera
discovery is challenging due to the lack of common programming
interfaces of the websites hosting network cameras. Different brands
of network cameras have different programming interfaces. Different
institutions organize the data in different ways.  Such heterogeneity
hinders the usability of the real-time data in emergencies. In other
words, network camera data is not readily indexed. For example, there
is no easy way to generate images from all the live public network
cameras in New York City. A web search will yield websites that point
to camera data in New York City. But the data is spread across many
websites, and it is not clear how to easily aggregate images from
relevant cameras.  To solve this problem, the CAM$^{2}$ team is (1)
building a web-crawler to work with many different website interfaces
and (2) building a database to provide a uniform interface via a
RESTful API.  The current version of the RESTful API has been released
and the Purdue continues discovering network cameras.

\subsection{Contextual Information as Weak Labels}\label{subsec:weakLabels}

The proper estimation of meta-data related to each network camera
provides useful functionality in the future. Location and time of day
provides useful information for automatic dataset
augmentation. Information such as location and time can be called {\it
contextual information} of the image/video data.  As an example, a camera deployed
on a busy highway is unlikely to see rhinos or buffaloes.  If such an
animal does appear on the highway, this unusual event is likely
reported by news (also can be looked up by time and location).  In
contrast, a network camera watching a waterfall in a national park
should not see semi-trucks.  Time also provides contextual information
about the visual data.  The streets in New York City are usually
filled by vehicles. On rare occasions (such as a parade shown in
Figures~\ref{fig:sample_images} (e)-(g)), the streets are filled by people.  Thus, 
this network cameras data can provide \emph{almost correct} labels by simply
assuming there exists vehicles in the data, modifying the label with cross-referenced news reports (of a parade)
and other anomaly detection can form a type of
\emph{weak
  supervision}~\cite{Ratner:2018:SMW:3209889.3209898}.

Weak supervision refers to using labels that can be (1)
automatically generated using partially true rules
(c.f. examples above), (2) utilizing related ground-truth labels that
are not for exactly the same task, (3) boosting, (4) hand labeling
with unreliable, non-expert annotators. Current research demonstrates
how different types of weak supervision can be used to improve the
accuracy of machine learning
models~\cite{Ratner:2018:SMW:3209889.3209898}.  Contextual information
provides weak labels similar to (1), and we suspect future work will
also improve model accuracy for image and video tasks, such as
classification and object detection.

This contextual information can be easily derived if a GPS (Global
Positioning System) receiver is included in every camera that is
deployed outdoors. GPS receivers are already in every mobile phone and
it is expected that all future network cameras will also be equipped with
GPS receivers.  Time and location may be referenced by sunlight
and sun location. Location can be even further refined by significant
landmarks. For cameras deployed indoors, methods also exist for
positioning them~\cite{indoorpositioning}.

Other contextual information such as indoor/outdoor and urban/rural
can be derived from a set of images using a variety of available
computer vision methods. If needed, a
new dataset of contextual information can be created. By training a
computer vision method on the dataset, and contextual information can
be automatically generated. While this is only an approximate
solution, it is feasible that this will be sufficient for weak labels.

\subsection{Network Camera Data is Distinct}\label{subsec:dataDistinct_opp}

As described in~\cite{GauenIRI}, commonly used
datasets are distinct from each other. For example, the images used in
ImageNet~\cite{ImageNet} can be distinguished from images used in
COCO~\cite{COCO}. Furthermore for object detection tasks, labelled
objects are more centrally concentrated for ImageNet than for COCO.
This difference is a result of the different data distributions.  Existing
computer vision solutions tend to focus on developing accurate models
for a small number of data
distributions.  Even when
models are compared across many different datasets, the solutions'
applicability beyond these datasets is unclear.  The testing error of
recently developed models can be overly optimistic even for samples from
the same data distribution~\cite{recht2018cifar10}.  Since the visual
data from network cameras may be considered {\it distinct} from other
datasets, the simple re-purposing of a model's weights
from a similar task may be insufficient.  Instead,
more sophitocated {\it transfer learning}
methods may be required to mitigate the differences among
datasets. Additional evidence would be needed to demonstrate that such
techniques can handle the wide range of visual data from thousands of
network cameras. Future work can investigate the degree to which
models trained on available data can be transferred, and the best
method for transferring the model's
information. If needed, this
may require an expansion of the existing CAM$^{2}$ dataset for each
specific application~\cite{GauenIRI}.

\subsection{Improving Models for Emergency Response}\label{subsec:emerg_opp}

The data seen during emergency events is uncommon. Thus, the accuracy
of machine learning models to respond to emergencies is likely
poor. We propose to three methods to improve machine learning in the
event of an emergency. (1) Periodically record data before a distaster
for an anomoly detection system. (2) Connect CAM$^2$ to a
infrastructure for crowd-sourcing to gather labelled data on short
notice from locations known to have an impending emergency
situtaion. For example, crowd-source was used in the Haiti earthquake
of 2010~\cite{Zook2010Haiti}. It is
conceivable to create a similar infrastructure for images and video in
an emergency. (3) Network cameras need to have a uniform interface for
easy access in emergency situtations, given the proper privacy and
legal constraints~\cite{Lu:2017:PPO:3123266.3133335}.

\subsection{Dataset Distinctiveness for Active Learning}\label{subsec:distinct_opp}

Finding the right subset of data to label is a general question in
machine learning, especially during an emergency when time is short.
The problem is also applicable to non-emergency scenarios when the
cost of labelling data is the constraint, rather than time.

As network camera data is distinct, we wish to further investigate if
dataset distinctiveness can be used to improve active learning
methods. In this paper, active learning is defined as following: Given
a large number of unlabeled data, we must identify the right subset
of data to label. A general framework for active learning methods is
often given as the balance between
measures of (1) how “representative” (or typical) the sample of data
is relative to the true data distribution and (2) maximizing variance
reduction of the model (equivalently, minimizing the true risk).
Often, the product of the two measures is used to identify the best
samples. With an input-output pair, this can be thought of as (1)
modeling only the data and (2) modeling only the conditional
distribution. Multiplying them together can be thought of as modeling
the unnormalized posterior.  Since network camera data is distinct
from the existing datasets, the use of this new data source may
improve the current active learning methods.

\subsection{Adaptive and Programmable Network Cameras}\label{subsec:adapt_opp}

Another improvement is to make network cameras ``self aware'' of the
context being seen, so as to automatically execute relevant
programs.  It may be possible for stationary cameras to determine the
visual information being captured and install/execute computer vision
programs specialized for the content. Moreover, network cameras may
need to be reprogrammed in emergencies.  The street cameras in
Figures~\ref{fig:sample_images} (e)-(g) may be specialized for detecting congestion
and accidents in normal conditions.  During a parade, the cameras may
need to be reprogrammed to search for a lost child~\cite{Satyanarayanan:2010:MCN:1810931.1810936}.

\section{Conclusion}

Network cameras provide rich information about the world. The visual data has many applications,
including real-time emergency response and discovery of long-term trends.
This paper presents a software infrastructure called CAM$^2$ (Continuous
Analysis of Many Cameras) constructed at Purdue University for
acquiring and analyzing data from network cameras.
The paper suggests many opportunities using the data and challenges to be conquered.

\section*{Acknowledgments}

This project is supported in part by Facebook, Intel,
the National Science Foundation OISE-1427808, IIP-1530914, OAC-1535108, OAC-1747694. Any
opinions, findings, and conclusions or recommendations
expressed in this material are those of the authors and do
 not necessarily reflect the views of the sponsors.

The authors would like to thank Amazon and Microsoft for providing the
cloud infrastructure, and the organizations that provide the
camera data. Many Purdue students taking VIP (vertically
 integrated projects) have contributed to the development
 of the software components used in CAM$^2$.

\section*{Authors}

Yung-Hsiang Lu is a professor in the School of Electrical
and Computer Engineering of Purdue University. His research
interests include computer vision, distributed sytems,
and low-power systems. He received Ph.D. from Stanford University. He is a senior member of the IEEE and a Distinguished Scientist of the ACM. Contact him at yunglu@purdue.edu.

George K. Thiruvathukal is a professor in the Department of
Computer Science of Loyola University Chicago
and visiting faculty at Argonne National Laboratory. His research interests include parallel computing, software engineering, and programming languages. He received Ph.D. from Illinois Institute of Technology. Contact him at gkt@cs.luc.edu.

Ahmed S. Kaseb is an assistant professor in the Faculty of Engineering at Cairo University. His research interests include cloud computing, big data analytics, and computer vision.
He received Ph.D. from Purdue University. Contact him at
akaseb@eng.cu.edu.eg.

Kent Gauen is a doctoral student in the School of Electrical
and Computer Engineering of Purdue University. His research
interests include computer vision and machine learning. He recevied BSECE from Purdue University.
Contact him at gauenk@purdue.edu.

Damini Rijhwani is an undergraduate student in the School of Electrical and Computer Engineering of Purdue University. Her research interests include computer vision and machine learning.
Contact her at drijhwan@purdue.edu.

Ryan Dailey is a master student  in the School of Electrical and Computer Engineering of Purdue University. His research interests include distributed systems and data mining.  He received BSECE from Purdue University. Contact him
at dailey1@purdue.edu.

Deeptanshu	Malik is a software engineer at Microsoft. He received BSECE from Purdue University. Contact him at malikd@purdue.edu.

Yutong	Huang is a master student  in the School of Electrical and Computer Engineering of Purdue University. His research interests include operating systems and distributed memory management.  He received BSECE from Purdue University. Contact him at huang637@purdue.edu.

Sarah Aghajanzadeh is a master student  in the School of Electrical and Computer Engineering of Purdue University. Her research interests include computer vision and machine learning. Contact her at saghajan@purdue.edu.

Minghao Guo is a master student in Data Science of  University of Illinois at Urbana Champaign. Her research interests include 
data mining, visualization, and statistical data analysis.
She received BSCS from Purdue University. Contact her at 
minaaa1003@gmail.com.

\bibliographystyle{unsrt}
\bibliography{Master}
\end{document}